%% file: manuscript.tex
\documentclass{article} 
\usepackage{iclr2025_conference,times}
\usepackage{graphicx}
\usepackage{soul}
\usepackage{tabularx}

\input{math_commands.tex}

\usepackage{hyperref}
\usepackage{url}

\title{MoE Lens - An Expert Is All You Need}

\author{
Marmik Chaudhari$^{1}$\thanks{Equal contributions.\\Our code is available at: \url{https://github.com/MarmikChaudhari/moe-interp/tree/moe-lens}} ,
Idhant Gulati$^{1*}$,
Nishkal Hundia$^{2*}$,
Pranav Karra$^{1*}$,
Shivam Raval$^{3}$ \\
\\
$^1$The Pennsylvania State University, University Park, PA 16802 USA \\
$^2$University of Maryland, College Park, MD 20742 USA \\
$^3$Harvard University, Cambridge, MA 02138 USA \\
\\
\texttt{\{\protect\href{mailto:marmik@psu.edu}{marmik}, \protect\href{mailto:idhant@psu.edu}{idhant}\}@berkeley.edu, \protect\href{mailto:pranavkarra@psu.edu}{pranavkarra@psu.edu}} \\
\texttt{\protect\href{mailto:nhundia@umd.edu}{nhundia@umd.edu}, \protect\href{mailto:sraval@harvard.edu}{sraval@harvard.edu}}
}

\iclrfinalcopy 
\begin{document}

\maketitle


\begin{abstract}
Mixture of Experts (MoE) models enable parameter-efficient scaling through sparse expert activations, yet optimizing their inference and memory costs remains challenging due to limited understanding of their specialization behavior. We present a systematic analysis of expert specialization in MoEs through two complementary approaches: domain-specific routing patterns and an early decoding framework that tracks expert contributions to output representations. Our analysis of the DeepSeekMoE model reveals that despite having 64 routed experts with 6 active for each layer's computation, the model predominantly relies on a few specialized experts, with the top-weighted expert's output closely approximating the full ensemble prediction. We quantitatively validate these findings through a systematic analysis of the token routing distribution, demonstrating that very few experts handle over 50\% of routing decisions across different specialized domains. Hidden state similarity between single and ensemble experts for every layer is extremely high, with some layers having cosine similarity as high as 0.95 and perplexity increasing by only 5\% when using a single expert across all three domains. Our results indicate that Mixture of Experts models exhibit concentrated expertise highlighting potential opportunities for inference optimization through targeted expert pruning while maintaining model performance and opening avenues towards studying localization of learned knowledge in these models.
\end{abstract}

\section{Introduction}
\label{gen_inst}

Mixture of Experts (MoE) \citep{outrageously-large-neural-nets} models offer an efficient way to scale large language models by activating only a subset of model parameters for each input. 
However, MoE architectures face challenges spanning from training complexity and load balancing to routing inefficiencies and memory constraints \citep{liu2025surveyinferenceoptimizationtechniques}, with many issues arising from how inputs are routed to experts and how specialization emerges. Recent architectures like DeepSeekMoE \citep{deepseek-moe} have improved expert specialization and load balancing \citep{aux-load-balancing-moe}, demonstrating progress in mitigating some of these challenges but fundamental questions about the expert behavior like specialization and knowledge redundancy in a MoE still remain unanswered. 

\begin{figure}[h]
    \center
    \includegraphics[width=0.97\textwidth]{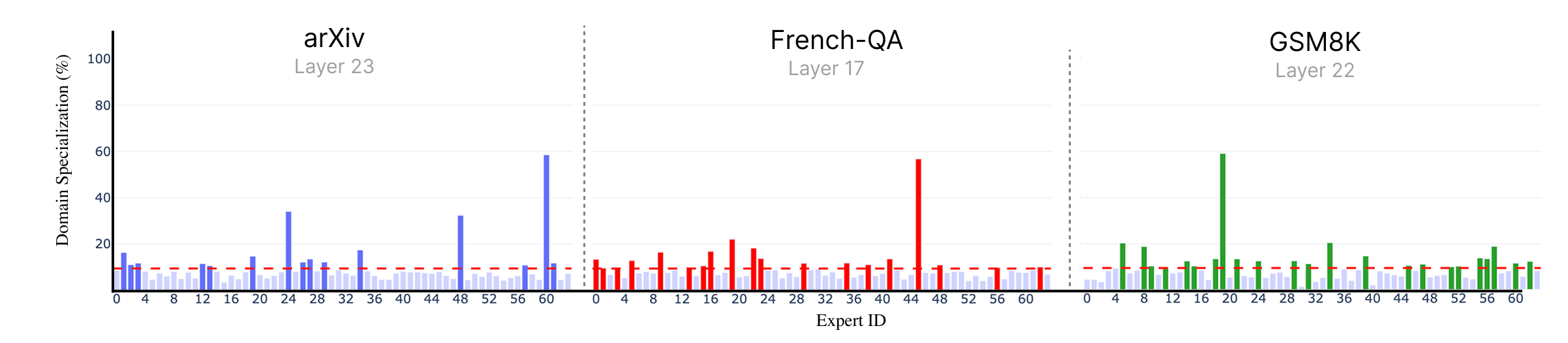}
    \caption{\textbf{Expert Specialization} in DeepSeekMoE. We visualize the distribution of tokens that are routed to an expert for our English, French-QA, and GSM8K datasets. The y-axis shows the routing percentage per expert, with the red dashed line indicating a uniform routing baseline ($\approx$ 9.4\%). See Appendix \ref{subsec:expert-specialization} for extended plots for other models and layers.}
    \label{fig:expert-specialization}
\end{figure}


 
Expert specialization is an emergent property first observed in vision models like AlexNet \citep{NIPS2012_c399862d} and InceptionV1 \citep{szegedy2014goingdeeperconvolutions}, where different network branches develop specialized feature representations. Language Models like Monet \citep{monet} have demonstrated that drastically scaling up the number of experts, individual experts develop specialized classes like chemical compounds, electromagnetism and diseases. MoE architectures with relatively fewer experts exhibit knowledge redundancy \citep{multilinear-moe} where a few experts cover the same diverse, unrelated concepts. This raises the question of how to identify and analyze the experts in a MoE that develop into monosemantic units, each specializing in distinct linguistic or computational domains, and how to leverage specialized experts to reduce the inference latency while preserving the knowledge and reasoning capabilities learned by the model during pre-training.

In this paper, we: (1) investigate expert specialization behavior across distinct domains by analyzing routing distributions and identifying domain-specialized experts across three data domains, (2) use an early-decoding strategy to interpret how individual experts contribute to residual stream representations at each layer, and find that a single expert is sufficient to converge to output representations in next-token prediction tasks, and (3) validate our findings through quantitative measures like cosine similarity and perplexity. Our empirical results reveal that while certain MoEs exhibit some domain specialization, they primarily rely on a small set of experts with other experts providing minimal contributions to the final predictions building towards an interpretable pruning approach that preserves the next token prediction accuracy while making the model more sparse.

\section{Background}



\textbf{MoE Layer.} In a Mixture-of-Experts (MoEs) architecture \citep{olmoe, deepseek-moe, open-moe}, the Feed-Forward Network (FFN) in a Transformer is substituted with MoE layers at specified intervals. The MoE layer consists of set of $n$ experts $E_1,...,E_n$, each structurally identical to a standard FFN, and a learned routing network, $r$, which assigns routing probabilities to all $n$ experts for each input token, $x$. The output of the $\ell$-th MoE layer for the $t$-th token, $\mathbf{h}_t^\ell$, is a weighted sum of the expert outputs scaled by  their corresponding routing probability across all chosen Top-$k$ experts, ${E_i}$ where ${i \in \text{top-}k}$. Let $\mathbf{u}_t^\ell $ denote the hidden state of the $t$-th token after the $l$-th attention module (post-attention residual stream). Mathematically,

\begin{equation}
\mathbf{h}_t^\ell = \sum_{i=1}^{\text{Top-}k(r_{t}(x))} (r_{i,t} E_i(x)) + \mathbf{u}_t^\ell,
\end{equation}

To analyze how experts specialize in processing different types of inputs, we define \textbf{expert specialization} as described in \cite{olmoe} to be the fraction of tokens from a particular domain $D$ for which expert $E_i$ is selected as one of the top-$k$ experts. 

\begin{equation}
\text{Expert specialization}(E_i, D) = \frac{N_{E_i,D}^{(k)}}{N_D},
\label{eq:expert-specialization}
\end{equation}

where $N_{E_i,D}^{(k)}$ is the number of tokens from domain $D$ for which $E_i$ is among the top-$k$ selected experts, and $N_D$ is the total number of tokens from domain $D$. We consider an expert to be specialized in $D$ if it processes significantly more tokens than the uniform routing baseline of 6/64 $\approx$ 9.4\%.


\textbf{Shared Experts}. Every input token is routed to the Shared Expert in DeepSeekMoE and the output of the Shared Expert, $E_S$ is added to the output hidden state, $\mathbf{h}_t^\ell$.  We primarily focus on the routed experts since the shared experts are dedicated to capturing common knowledge across varying contexts \citep{deepseek-moe}

\begin{figure}[h]
    \center
    \includegraphics[width=0.98\textwidth]{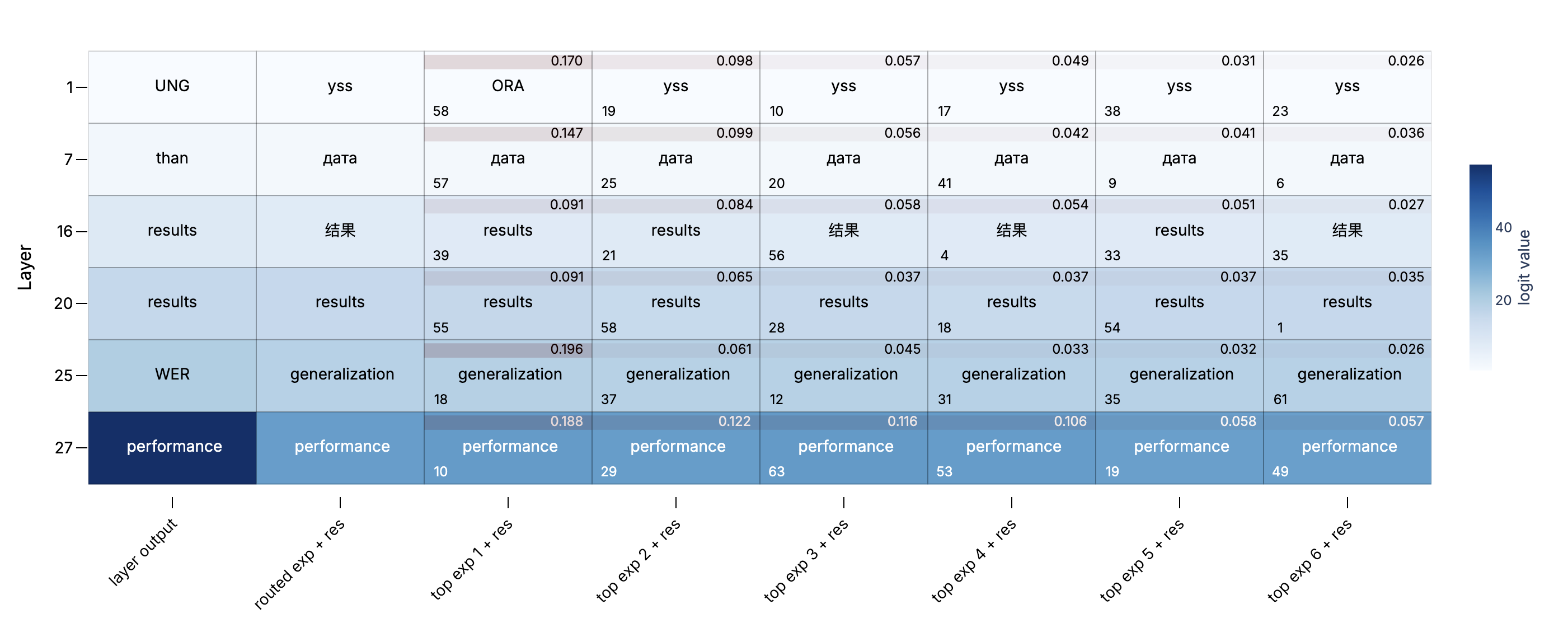}
     \caption{An example of early decoding using \textbf{LogitLens} for DeepSeekMoE on an example input: ``When datasets are sufficiently large, increasing the capacity (number of parameters) of neural networks can give much better prediction''. Each cell shows the top-1 token prediction after the final token ``these'' across layers (rows) for layer output, routed experts + residual stream for various top-$k$ values. Color intensity indicates prediction confidence. The expert index is denoted by the lower-left subscript number and the top-right superscript indicates expert weight. See Appendix \ref{subsec:logit-lens} for other domains.}
    \label{fig:logit-lens}
\end{figure}
\textbf{Early decoding using Logit Lens.} Early decoding \citep{schuster2022confidentadaptivelanguagemodeling} is the analysis of a model's intermediate predictions before reaching the final layer. The LogitLens \citep{logit-lens} is an early decoding technique that directly decodes the hidden states at any intermediate layer $\ell$ for $t$-th token, $\mathbf{h}_t^\ell$, using the model's pretrained unembedding matrix, $W_U$. The resulting distribution of logits roughly converges toward the model's final prediction across layers, offering a window into how the model progressively refines its predictions.
\begin{equation}
	\text{LogitLens}({\mathbf{h}^\ell_t}) = \text{LayerNorm}({\mathbf{h}^\ell_t})W_U
	\label{eq:logit-lens}
\end{equation}
To understand how expert, $E_i$, at a particular layer $\ell$ contributes to the final output representation, we further extend the LogitLens by adding post-attention residual stream for an expert $\mathbf{u}_t^\ell$ and then projecting it to the vocabulary space. The residual stream is analogous to a communication channel \citep{elhage2021mathematical} through which the experts incrementally refine the hidden state, ${\mathbf{h}^\ell_t}$. Each expert's output can be interpreted as a targeted modification to specific subspaces of the representation. By using extended LogitLens \citep{belrose2023elicitinglatentpredictionstransformers} , we observe how individual experts update the prediction distribution by writing their specialized knowledge into the residual stream.

\begin{equation}
	\text{LogitLens}^{ext}({\mathbf{h}^\ell_t}) = \text{LayerNorm}({\mathbf{h}^\ell_t} + \mathbf{u}_t^\ell)W_U 
	\label{eq:extended-logit-lens}
\end{equation}

\textbf{Notation.} For layer $\ell$, we represent the hidden state of the top-weighted expert (top-$k=1$) combined with the residual output as $\mathbf{H_{t}^{\ell_1}}$ and the hidden state of the top-$k=6$ weighted expert combined with the residual output as $\mathbf{H_{t}^{\ell_6}}$ .

\section{Experiments}


\begin{figure}[h]
    \center
    \includegraphics[width=0.95\textwidth]{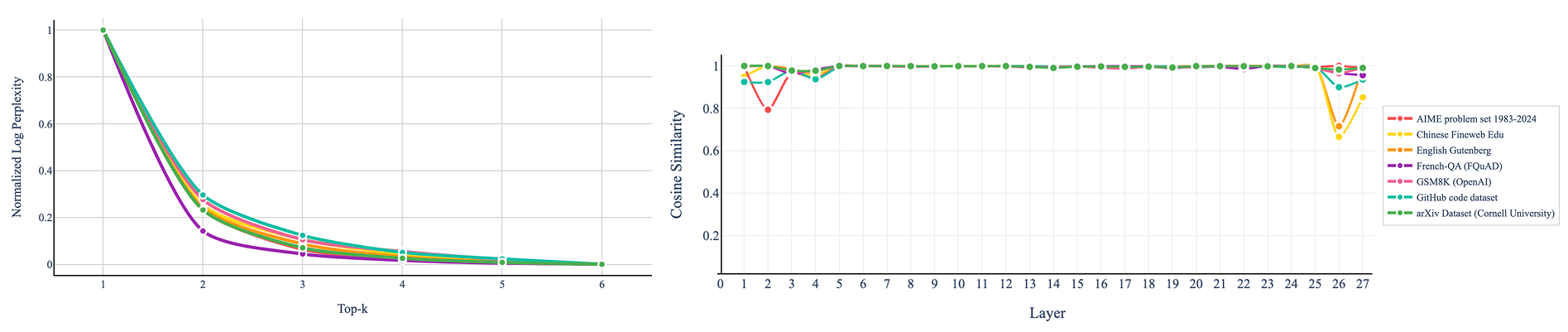}
    \caption{(Left) Normalized log perplexity across different values of top-k experts for various domains for next-token prediction task. (Right) Cosine similarity between the hidden states of $\mathbf{H_{t}^{\ell_1}}$ and $\mathbf{H_{t}^{\ell_6}}$ across all 27 layers shows consistently high alignment.}
    \label{fig:cosin-sim-ppx}
\end{figure}

\textbf{Model.} For all experiments, we use DeepSeekMoE \citep{deepseek-moe} with 2 shared + 64 routed experts with top-$k=6$. The model is pretrained on cross entropy loss combined with expert-level and device-level balance loss to prevent a router collapse. We also extended our experiments by utilizing OLMoE by Allen AI \citep{olmoe}. See Appendix \ref{sec:appendix}.


\textbf{Datasets.} We use curated subsets of seven datasets for our experiments: (1) GitHub code dataset subset of Paloma \citep{paloma}, (2) subset of Gutenberg English Dataset \citep{gutenberg-eng}, a dataset of English books, (3) French-QA subset of French Question Answering Dataset (FQuAD) \citep{FQuAD}, a dataset on native French reading comprehension, (4) AIME Problem Sets from 1983 to 2024 \citep{aime_1983_2024}, (5) Chinese Fineweb Edu \citep{chinese-fineweb}, a high-quality Chinese pre-training corpus, (6) arXiv Dataset \citep{arxiv-dataset}, a collection of pre-print scholarly articles gathered from various scientific fields, and (7) GSM8K \citep{gsm8k}, a grade school math word problems dataset created by OpenAI.

\textbf{Experiments.} Our empirical investigation consists of three experiments: (i) For each input prompt in domain $D$ $\in$ \{English, Code, French\}, we compute Expert specialization $(E_i, D)$ as defined in \eqref{eq:expert-specialization}. (ii) We apply extended LogitLens (\eqref{eq:extended-logit-lens}) to analyze components within MoE layers. We project three distinct hidden states to logits: (1) individual expert outputs, $E_i$, (2) the weighted sum of the top-$k = 6$ expert outputs, and (3) final layer outputs. For each layer $\ell$, we compare the layer output, $\mathbf{h}^\ell_t$ , and the top-weighted expert output combined with the residual output, $\mathbf{H_{t}^{\ell_1}}$, for next-token prediction tasks from each domain, $D$. (iii) We compute the cosine similarity between the hidden state of the top weighted expert's output (expert + residual), $\mathbf{H_{t}^{\ell_1}}$, and the hidden state of the top-$k=6$ weighted expert outputs combined with residual output, $\mathbf{H_{t}^{\ell_6}}$, across each domain dataset to assess the alignment between individual and combined expert contributions in the hidden space. We also analyze the perplexity of the next token prediction for each domain $D$ and how it changes with different number of active experts to observe whether using the single most-activated expert maintains comparable loss and prediction performance to using all 6 routed experts. 

\section{Results}

The expert specialization results in DeepSeekMoE, shown in \figref{fig:expert-specialization}, reveal two key patterns: First, only a small number of experts show strong specialization (significantly higher than uniform routing frequency) for any domain. Second, most experts demonstrate minimal domain-specific activity.

The extended LogitLens gives us evidence that solely projecting $\mathbf{H_{t}^{\ell_1}}$ across layers decode to roughly the same next token prediction as the output at the end of that layer, $\mathbf{h}^\ell_t$ . Furthermore, $\mathbf{H_{t}^{\ell_1}}$ has nearly identical next-token distribution as $\mathbf{H_{t}^{\ell_6}}$. 

We also observe very high cosine similarity between $\mathbf{H_{t}^{\ell_1}}$ and $\mathbf{H_{t}^{\ell_6}}$ across all layers and each domain, $D$ as shown in \figref{fig:cosin-sim-ppx} (right) indicating that the top-weighted expert is contributing the most in shaping final output representation whereas the contributions of other experts are minimal in the hidden space. Concretely, $\mathbf{H_{t}^{\ell_1}} \approx \mathbf{H_{t}^{\ell_6}}$. The perplexity moderately increases when reducing top-$k=6$ to $1$ as demonstrated in \figref{fig:cosin-sim-ppx} (right) which validates the claim that the top-weighted expert, when combined with the residual stream, produces representations closely aligned with the layer output.

\section{Conclusions and future work}
We empirically show that a single top-weighted expert, combined with the residual stream, closely approximates the full ensemble output for a layer across multiple data domains in DeepSeekMoE. Such a high degree of specialization suggests a potential for further sparsification during inference. By activating only the highest-weighted expert instead of all top-k experts and selectively pruning the non-essential experts,  computational costs and memory requirements can be significantly reduced while maintaining comparable model performance across a variety of tasks. 

We demonstrate these findings using DeepSeekMoE as a representative MoE architecture. Examining additional MoE variants such as OLMoE \citep{olmoe}, DeepSeek-V2 \citep{deepseekai2024deepseekv2strongeconomicalefficient}, and DeepSeek-VL2 \citep{wu2024deepseekvl2mixtureofexpertsvisionlanguagemodels} can lead to a broader understanding of specialization behavior in MoE models. While our analysis uses the LogitLens approach, extending this work by using TunedLens may provide more robust token decoding through layer-wise learned transformation between intermediate and prefinal layer representations.
Furthermore, these findings open several research directions: developing dynamic expert selection strategies that adapt to input complexity, analyzing the internal representation sparsity of individual experts to localize factual knowledge, and examining specialization patterns across different MoE architectures to inform future training objectives.

 
\bibliography{iclr2025_conference.bib}
\bibliographystyle{iclr2025_conference}
\newpage
\appendix
\section{Appendix}
\label{sec:appendix}

\subsection{Expert Specialization}
\label{subsec:expert-specialization}
\subsubsection{DeepSeekMoE}

\begin{figure}[ht!]
    \center
    \includegraphics[width=0.85\textwidth]{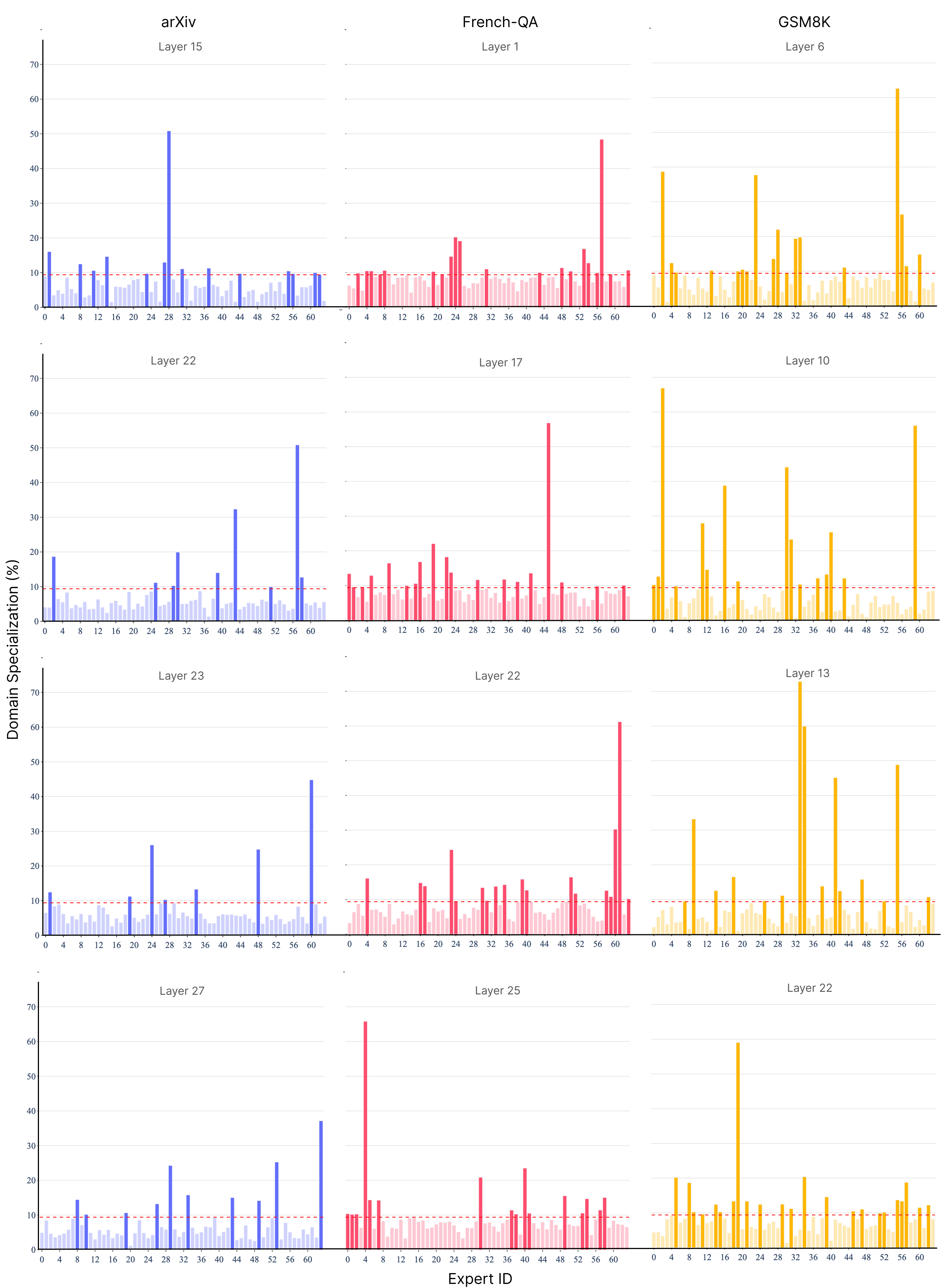}
    \caption{\textbf{Expert Specialization (Part 1)} of DeepSeekMoE for various layers. We visualize how frequently tokens from different domains are routed to the 64 experts using top-$k=6$ routing. The y-axis shows routing percentage per expert, with the red dashed line indicating uniform routing baseline ($\approx$ 9.4\%).}
    \label{fig:expert-specialization-1}
\end{figure}
\begin{figure}[ht!]
    \center
    \includegraphics[width=0.85\textwidth]{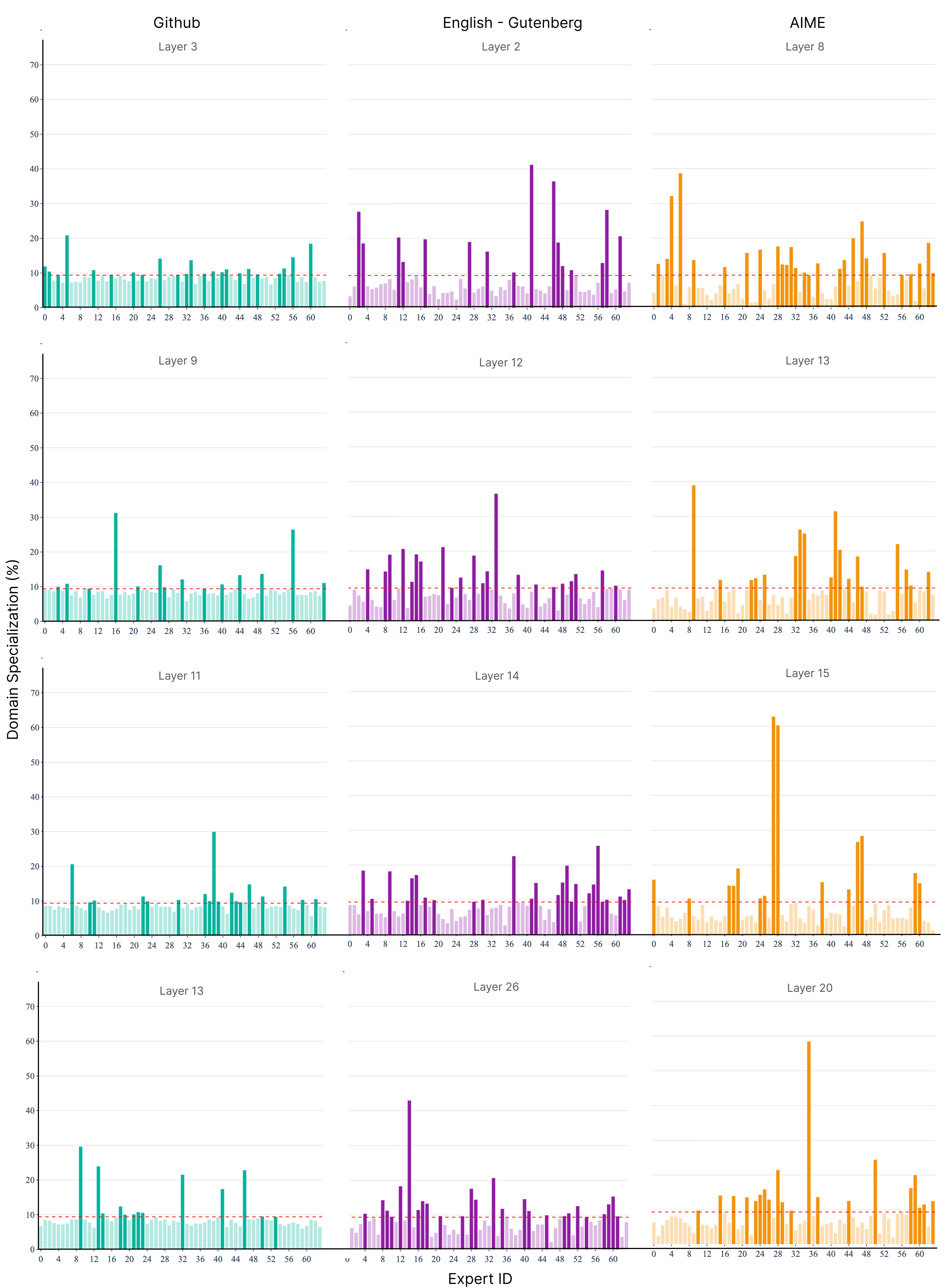}
    \caption{\textbf{Expert Specialization (Part 2)} of DeepSeekMoE for various layers. We visualize how frequently tokens from different domains are routed to the 64 experts using top-$k=6$ routing. The y-axis shows routing percentage per expert, with the red dashed line indicating uniform routing baseline ($\approx$ 9.4\%).}
    \label{fig:expert-specialization-2}
\end{figure}

\clearpage
\subsubsection{Qwen 1.5 MoE}
\begin{figure}[ht!]
    \center
    \includegraphics[width=0.85\textwidth]{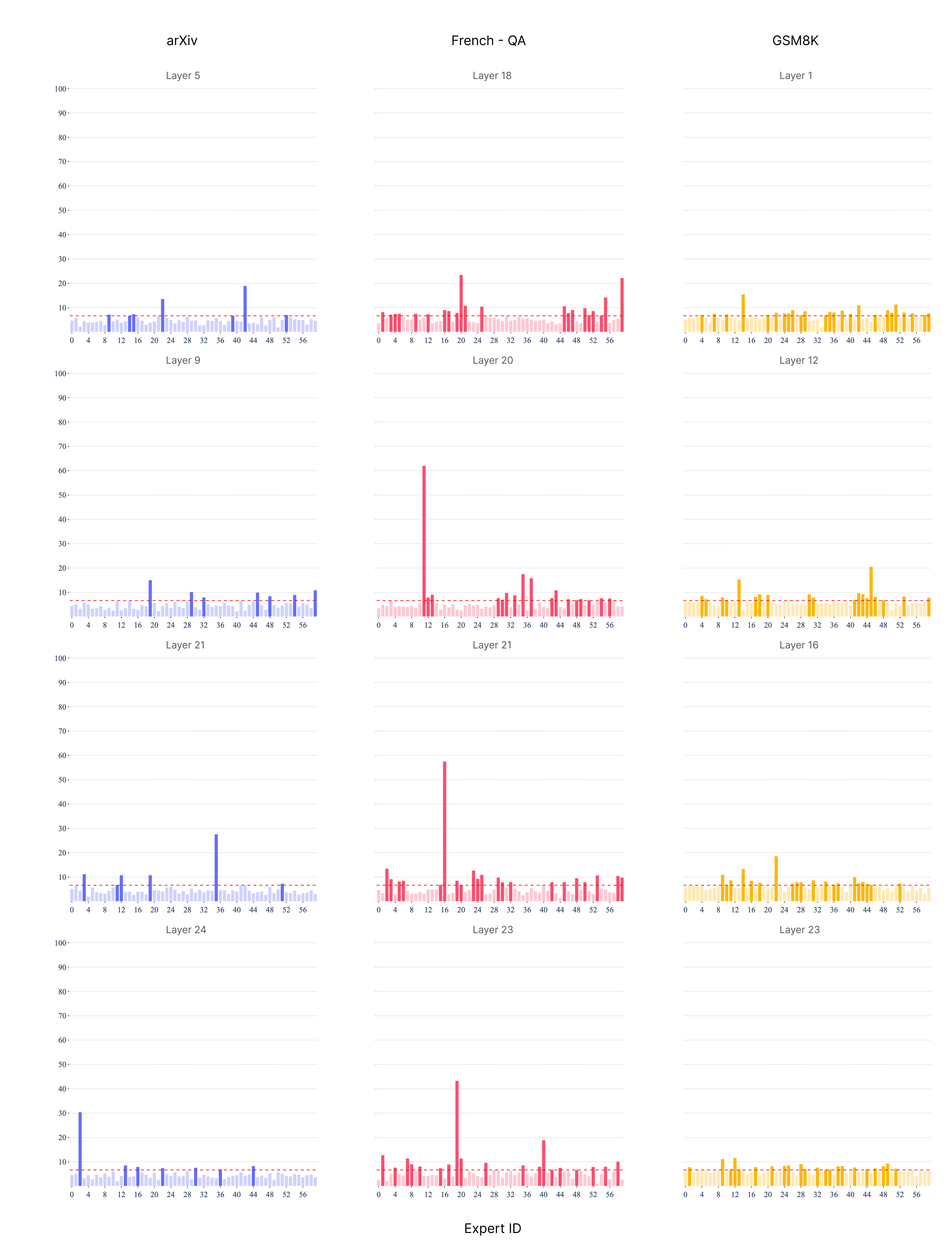}
    \caption{\textbf{Expert Specialization (Part 1)} of Qwen 1.5 MoE for various layers. We visualize how frequently tokens from different domains are routed to the 60 experts using top-$k=4$ routing. The y-axis shows routing percentage per expert, with the red dashed line indicating uniform routing baseline ($\approx$ 6.67\%).}
    \label{fig:expert-specialization-1-qwen}
\end{figure}
\begin{figure}[ht!]
    \center
    \includegraphics[width=0.85\textwidth]{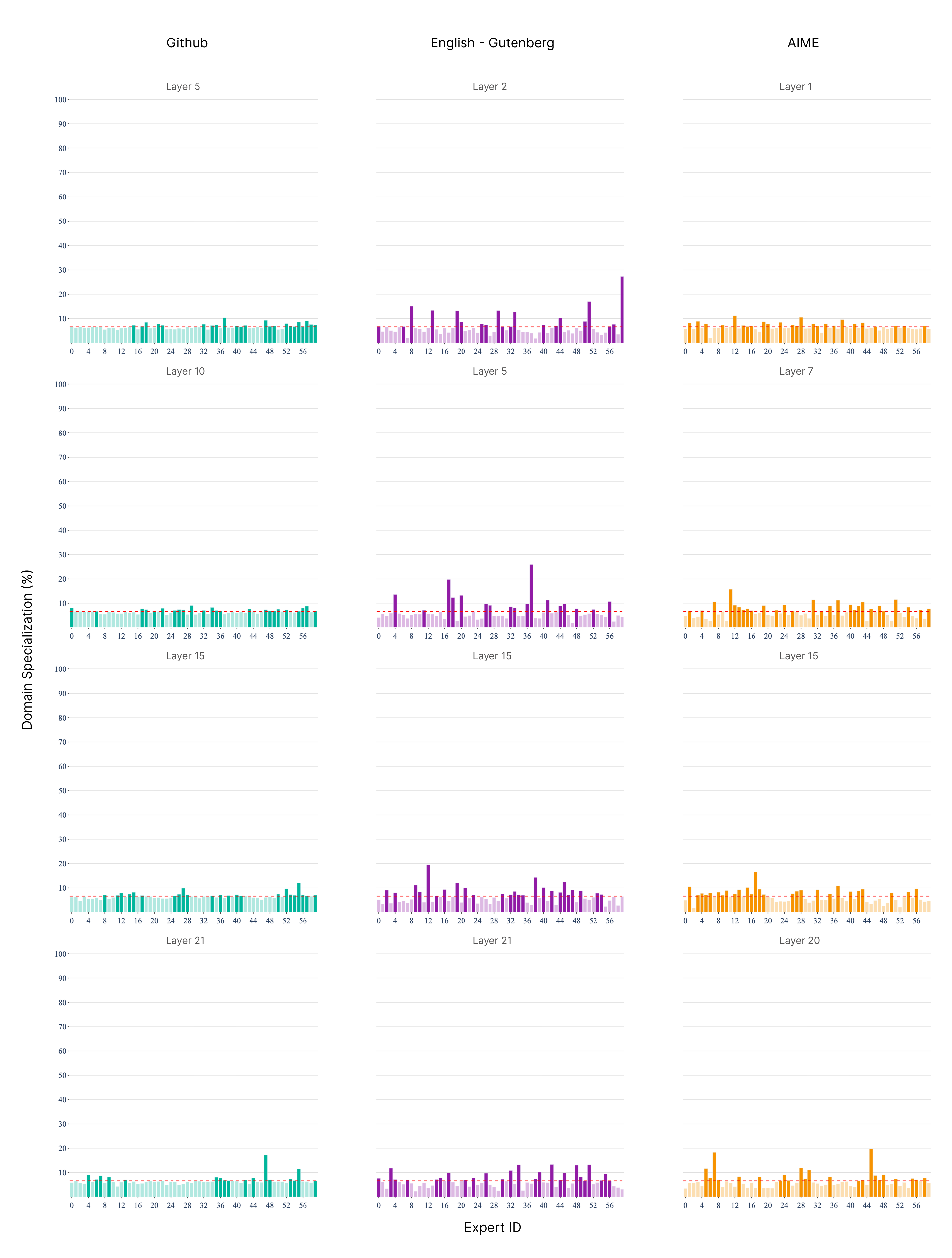}
    \caption{\textbf{Expert Specialization (Part 2)} of Qwen 1.5 MoE for various layers. We visualize how frequently tokens from different domains are routed to the 60 experts using top-$k=4$ routing. The y-axis shows routing percentage per expert, with the red dashed line indicating uniform routing baseline ($\approx$ 6.67\%).}
    \label{fig:expert-specialization-2-qwen}
\end{figure}

\clearpage
\subsubsection{OLMoE}
\begin{figure}[ht!]
    \center
    \includegraphics[width=0.85\textwidth]{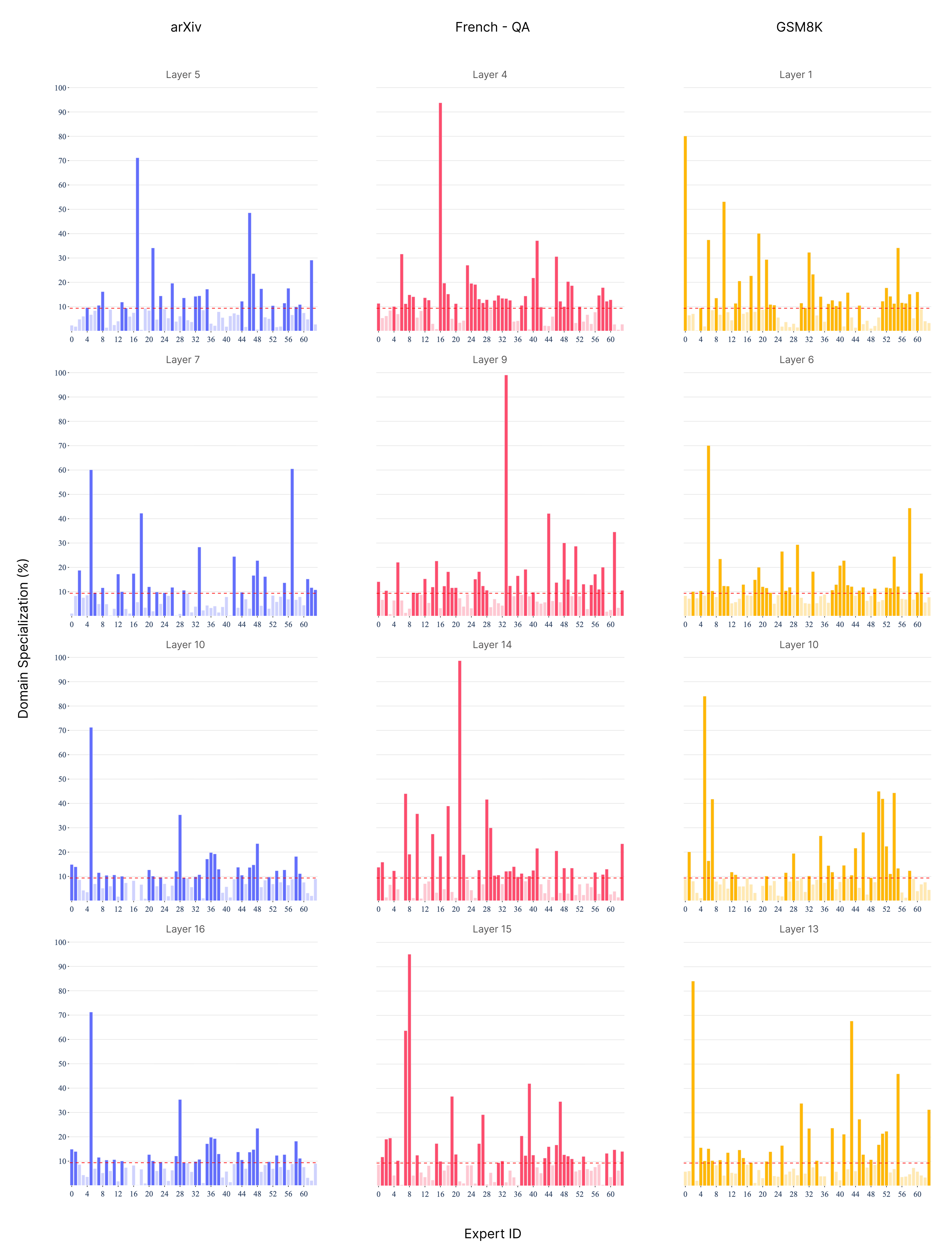}
    \caption{\textbf{Expert Specialization (Part 2)} of OLMoE for various layers. We visualize how frequently tokens from different domains are routed to the 64 experts using top-$k=8$ routing. The y-axis shows routing percentage per expert, with the red dashed line indicating uniform routing baseline ($\approx$ 12.5\%).}
    \label{fig:expert-specialization-1-olmoe}
\end{figure}
\begin{figure}[ht!]
    \center
    \includegraphics[width=0.85\textwidth]{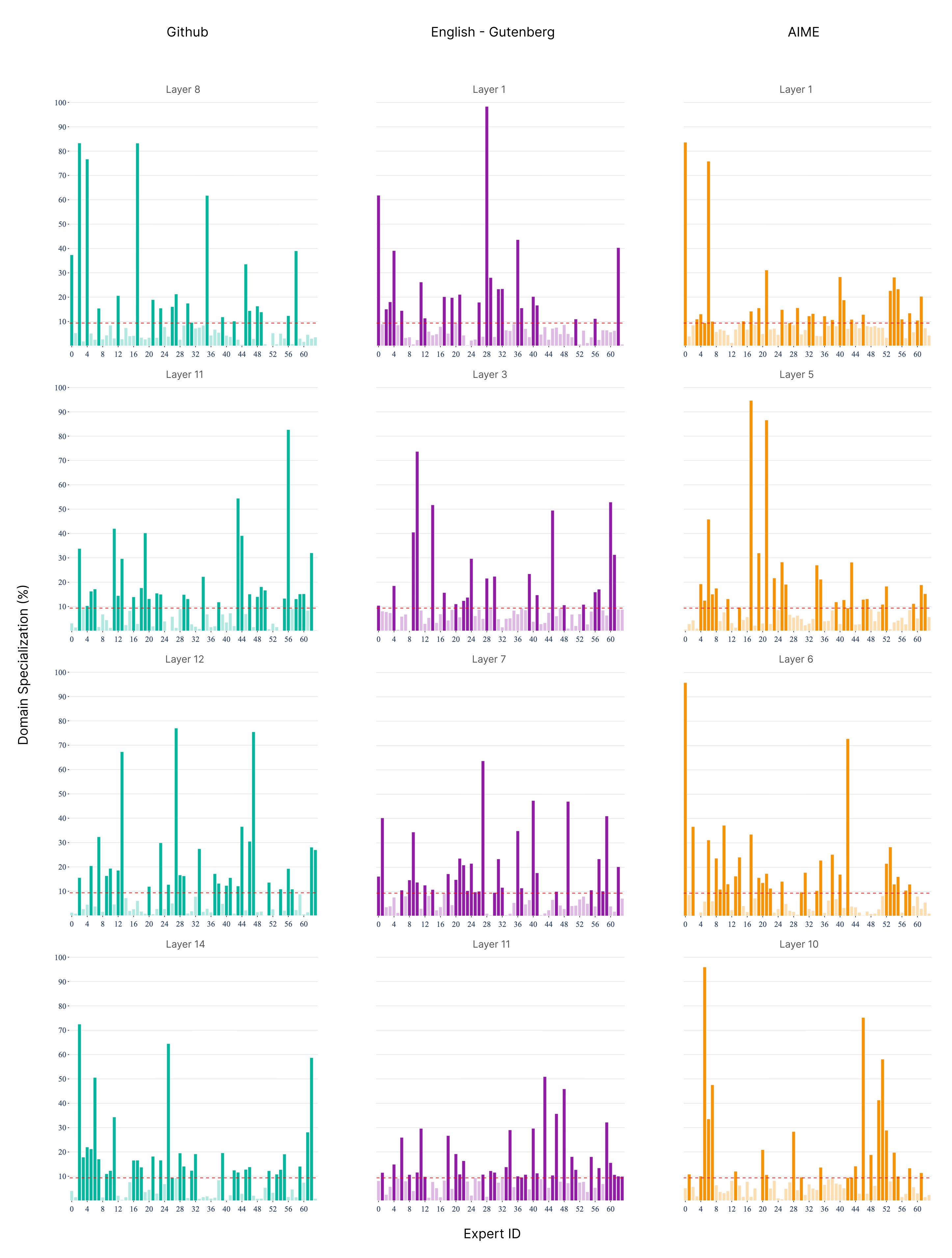}
    \caption{\textbf{Expert Specialization (Part 2)} of OLMoE for various layers. We visualize how frequently tokens from different domains are routed to the 66 experts using top-$k=8$ routing. The y-axis shows routing percentage per expert, with the red dashed line indicating uniform routing baseline ($\approx$ 12.5\%).}
    \label{fig:expert-specialization-2-olmoe}
\end{figure}

\clearpage
\subsection{Logit Lens}
\label{subsec:logit-lens}

\begin{figure}[ht!]
    \center
    \includegraphics[width=0.85\textwidth]{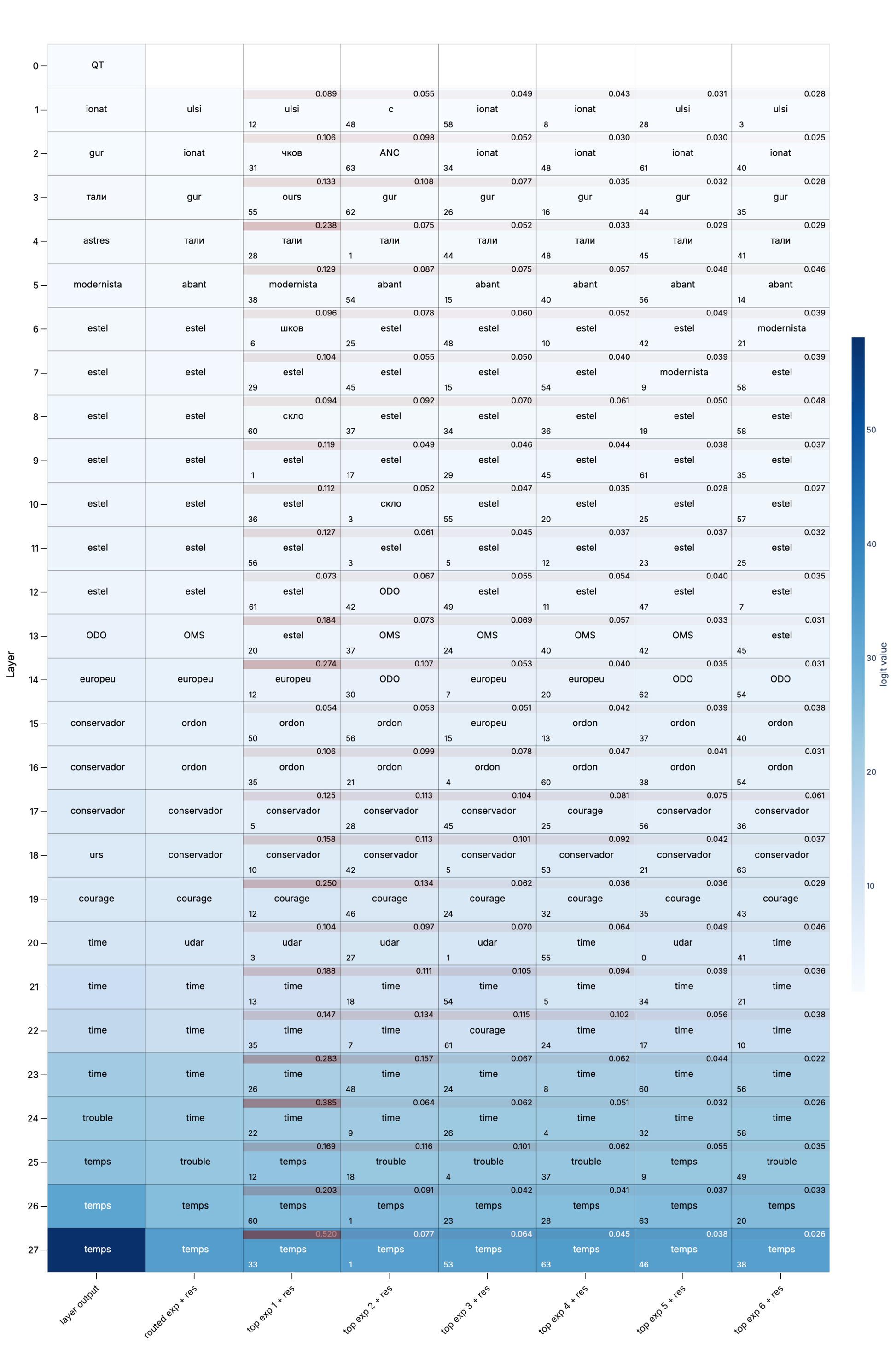}
    \caption{\textbf{LogitLens} visualization for DeepSeekMoE on the french input sequence: ``Dans le silence feutre de la nuit, les etoiles semblent murmurer d'anciens secrets a ceux qui prennent le''. Each cell shows the top-1 token prediction after ``le'' across layers (rows) for layer output, routed experts with residual stream for various top-$k$ values. Color intensity indicates prediction confidence. The lower-left subscript indicates expert indices and the top-right superscript indicates expert weight.}
    \label{fig:logit-lens-1}
\end{figure}

\begin{figure}[ht!]
    \center
    \includegraphics[width=0.85\textwidth]{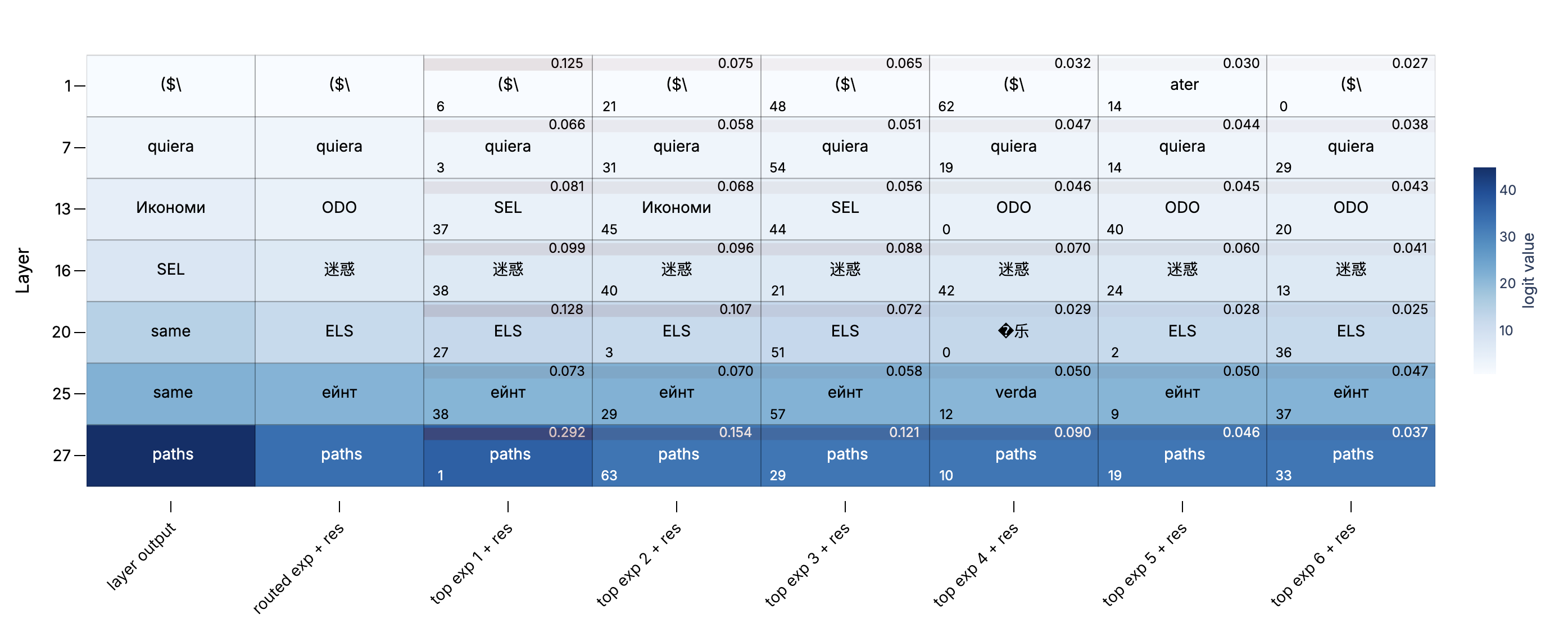}
    \caption{\textbf{LogitLens} visualization for DeepSeekMoE on the english input sequence: ``One might expect language modeling performance to depend on model architecture, the size of neural models, the computing power used to train them, and the data available for this''. Each cell shows the top-1 token prediction after ``this'' across layers (rows) for layer output, routed experts with residual stream for various top-$k$ values. Color intensity indicates prediction confidence. The lower-left subscript indicates expert indices and the top-right superscript indicates expert weight.}
    \label{fig:logit-lens-2}
\end{figure}


\end{document}

%% file: math_commands.tex
\usepackage{amsmath,amsfonts,bm}

\def\figref#1{figure~\ref{#1}}

\def\eqref#1{equation~\ref{#1}}

\def\1{\bm{1}}

\DeclareMathAlphabet{\mathsfit}{\encodingdefault}{\sfdefault}{m}{sl}
\SetMathAlphabet{\mathsfit}{bold}{\encodingdefault}{\sfdefault}{bx}{n}

